# EMRA-proxy: Enhancing Multi-Class Region Semantic Segmentation in Remote Sensing Images with Attention Proxy

Yichun Yu[1], Yuqing Lan[1,2], Zhihuan Xing[1*], Xiaoyi Yang[2], Tingyue Tang[1] and Dan Yu[3]

[1] School of Computer Science and Engineering, Beihang University, Beijing 100191, China
`yichunyu@buaa.edu.cn; Lanyuqing@buaa.edu.cn; xingzhhuan@buaa.edu.cn; zy2106143@buaa.edu.cn`
[2] School of Software, Beihang University, Beijing 100191, China
`xyyang001@buaa.edu.cn`
[3] China Standard Intelligent Security, Beijing 100097, China
`yu_dan@csisecurity.com.cn`

**Abstract.** Semantic segmentation is a highly challenging task in high-resolution remote sensing (HRRS) image due to the complex spatial layouts and significant appearance variations of multi-class objects. Convolutional Neural Networks (CNNs) have been widely employed as feature extractors for various visual tasks, owing to their excellent ability to extract local features. However, due to the inherent bias of convolutional operations, CNNs inevitably have limitations in modeling long-range dependencies. On the other hand, Transformers excel in capturing global representations but unfortunately overlook the details of local features and category features, and exhibit high computational and spatial complexity when dealing with high-resolution feature maps. Semantic segmentation has traditionally been modeled as predicting each point on a dense regular grid. In this work, we propose a novel and effective model, EMRA-proxy, which consists of two parts: homogeneous regions attention proxy (HRA-proxy) and Multi-class Attention proxy (MCA-proxy). The proposed EMRA-proxy model abandons the common Cartesian feature layout and operates purely at the region level. First, to capture contextual information within a region, we use Transformer to encode regions in a sequence-to-sequence manner by applying multiple layers of self-attention to region embeddings acting as proxies for specific regions. HRA-proxy then interprets the image into learnable surface subdivisions, each with flexible geometry and homogeneous semantics. It is performed by using a single linear classifier on top of the encoded region embeddings for prediction per region, thereby obtaining a homogeneous semantic mask feature map (HSMF-map). Then MCA-proxy learns the global class attention map (GCA-map) to make up for ViT's shortcomings in multi-class information extraction. Finally, HSMF-map and GCA-map are integrated to achieve high-precision multi-class remote sensing image segmentation. Extensive experiments on three public remote sensing datasets demonstrate the superiority of EMRA-proxy and indicate that the overall performance of our method outperforms state-of-the-art methods.

**Keywords:** Remote sensing Images, Semantic Segmentation, Homogeneous regions attention proxy, Multi-Class Attention Proxy.

# 1    Introduction

In recent years, with the rapid development of Earth observation and satellite technologies, an enormous volume of remote sensing data has emerged, leading to significant advancements in remote sensing image analysis. Semantic segmentation stands as one of the most crucial tasks in remote sensing, aiming to assign a category label to each pixel in an image. Semantic segmentation techniques have found widespread applications in environmental monitoring [1], smart agriculture [2], land cover detection [3], and urban planning [4].

The interpretation of high-resolution remote sensing (HRRS) images has long-standing challenges in image segmentation. Compared to natural images, HRRS images exhibit complex spatial layouts and diverse categories of objects [5]. As depicted in **Fig. 1**, objects of the same category display inconsistent category distributions across different scenes, with significant variations in shape and appearance. Ground features manifest at multiple scales with relatively intricate texture information, leading to high inter-class similarity and intra-class diversity, making it challenging to accurately delineate their boundaries. Additionally, the high resolution provides finer details for background samples and greater intra-class variance, requiring the extraction of more discriminative semantic features for precise segmentation. Over the past decade, many research methods have focused on pixel-wise classification for segmentation, employing traditional manual feature design and classifiers, which have yielded some encouraging results [6,7,8]. However, these methods are limited by the finite nature of feature extraction, leading to poor performance when dealing with large-scale, complex images, particularly severely restricting their application in remote sensing.

Deep learning methods have emerged as a powerful approach for learning high-level abstract features directly from raw data [9], overcoming the limitations of traditional methods that heavily rely on handcrafted features and have limited feature representation capabilities. Convolutional Neural Networks (CNNs) demonstrate strong proficiency in capturing local details, rendering them widely adopted across diverse visual recognition tasks [10,11,12]. Since the introduction of Fully Convolutional Networks (FCNs) [13], researchers have developed numerous segmentation methods based on CNNs, which offer significant advantages and greatly improve segmentation accuracy. U-Net [14], for instance, utilizes a symmetrical encoder-decoder structure and employs skip connections from shallow to deep layers to fuse multi-level features, recovering details lost due to stacked convolutional layers and downsampling operations. Various variants of U-Net have been proposed in subsequent studies. MAResU-Net [15] utilizes a linear attention mechanism to reshape the skip connections of the original U-Net network, thereby enhancing the utilization of detailed information and improving segmentation performance. Despite the excellent feature extraction capability of CNN-based methods, they are unable to effectively model long-range dependencies due to the inherent limitations of their structure. Consequently, many approaches tend to enhance contextual modeling by extending the receptive field of CNNs, such as increasing kernel size [16], employing atrous convolutions [17], introducing feature pyramids [18], or utilizing multi-scale modules [19]. While these methods enhance feature representation, CNNs still struggle to learn global semantic information due to the limited receptive field of convolutional kernels, which is crucial for dense prediction tasks.

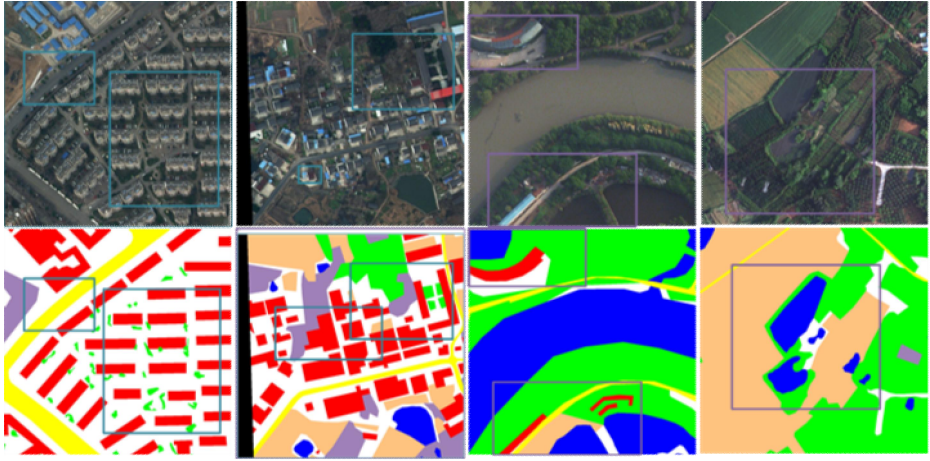

**Fig. 1.** In HRRS images, objects exhibit significant differences in shape, proportion, and spatial arrangement. In rural scenes, buildings often exhibit chaotic features, with narrow roads, interspersed with small ponds and trees. Conversely, urban scenes boast neatly arranged buildings, expansive roads, and a diminished presence of trees and ponds.

Transformer architecture-based semantic segmentation has gained popularity in recent years. The Transformer architecture [20] was initially devised for natural language processing, but has recently gained considerable interest in computer vision. It is well-suited for processing images with large receptive fields due to its ability to model long-range dependencies between input tokens via a self-attention mechanism. Particularly, the ViT [21] has demonstrated remarkable performance in image classification tasks by processing image segments directly as input. Several studies [22,23] utilizing ViT as a foundation for the semantic segmentation of Remote Sensing Images has produced remarkable results. ViT's attention mechanism facilitates the acquisition of more comprehensive contextual information. However, in current models, ViT only serves as a feature extractor for sequence-to-sequence encoders, extracting 2D coarse features similar to those of convolutional neural networks. Transformer-based models typically employ a single class token due to their limited understanding of the spatial relationships between objects, which makes it difficult to precisely localize various objects within a single image.

In this work, we explore a novel model for semantic segmentation that we believe is closer to its essence: we attempt to interpret multi-target images as a set of interrelated regions, where each region represents a group of adjacent pixels with homogeneous or class-consistent semantic information. As illustrated in **Fig. 2**, we present an EMRA-proxy model that aims to utilize homogeneous semantic information of image regions and object category information for segmenting HRRS images. The homogeneous regions attention proxy (HRA-proxy) is employed to identify homogeneous regions and object boundaries in the image during the early stages of the model. In addition, we use a Multi-class Attention proxy(MCA-proxy) to compensate for the insufficient extraction of multi-class information by ViT. In summary, our work provides three major contributions:

- We propose HRA-proxy, enabling operations on region embeddings across the network. These embeddings represent specific learnable regions, acting as proxies. Instead of pixel-wise predictions, we directly predict region embeddings for segmentation using a linear classifier. This approach effectively captures the homogeneous information of the image, enhancing the segmentation performance of complex geometric-shaped targets.
- We introduce MCA-proxy to address the limitation of ViT in extracting local features for multi-class targets, thereby compensating for inconsistent class distributions enabling the capture of a wider variety of class-specific feature information.
- We qualitatively and quantitatively evaluate the segmentation performance of EMRA-proxy on three challenging datasets, LoveDA, Potsdam and Vaihingen. Experimental results demonstrate a significant superiority over current methods.
- The remaining parts of this article are organized as follows.

In Section 2, a review of the literature on traditional image segmentation methods and deep learning-based semantic segmentation methods utilizing Convolutional Neural Networks and Transformer models is provided. In Section 3, we first introduce the overall architecture of EMRA-proxy. Following that, we provide a detailed explanation of the principles and implementation steps of HRA-proxy and MCA-proxy. Section 4 provides a detailed overview of our study's dataset and experimental setup. It covers the dataset's source, size, features, and the specifics of our experimental design, including model architecture, hyperparameters, training, and evaluation methods. We then present and analyze the results, offering insights into the performance and efficacy of our approach. The conclusion of this article will be drawn in Section 5.

## 2    Related work

**CNN-based Semantic Segmentation**. Semantic segmentation has always been an important task in remote sensing applications. Traditional methods generally rely on manual feature extraction, such as spectral information, image texture, and spatial features, followed by selecting feature classifiers to achieve segmentation. Due to high labor costs and inadequate feature representation, it is difficult to describe complex scenes accurately. In recent years, with the assistance of deep CNNs, various

FCN-based models have significantly improved the segmentation accuracy of remote-sensing images.

Addressing the challenge of complex scenes and diverse objects in remote sensing images commonly involves aggregating multi-scale contextual information. Zhao et al. [24] introduced attention mechanisms into the dilated spatial pyramid pooling module to achieve adaptive feature refinement and restored fine-grained features by fusing the pooling index map with high-level feature maps. Liu et al. [25] sequentially aggregated global-to-local context with low-level features to progressively refine target objects. Yue et al. [26] constructed a Tree-CNN block to fuse multi-scale feature maps, greatly enhancing discriminative power for confusable categories. Additionally, some studies focus on segmenting small objects. Zheng et al. [27] enhanced foreground feature recognition by learning foreground-related context from a foreground modeling perspective. Ma et al. [28] proposed a foreground-activated object representation framework to enhance the recognition of weak features in small objects. Peng et al. [16] encoded more accurate small-scale semantic information by selecting informative features and expanding the receptive field of low-level feature maps, significantly improving the segmentation accuracy of small targets. Another research direction is further improving results through attention mechanisms. Ding et al. [29] designed a patch attention module and attention embedding module to enhance feature representation and bridge spatial distribution differences between high-level and low-level features. Li et al. [30] proposed a boundary attention module to extract boundary information of objects from hierarchical feature aggregation and eliminate noise information from low-level features. However, the contextual information obtained by these models is still insufficient, and further improvement is needed for the segmentation accuracy of fine objects in HRRS images.

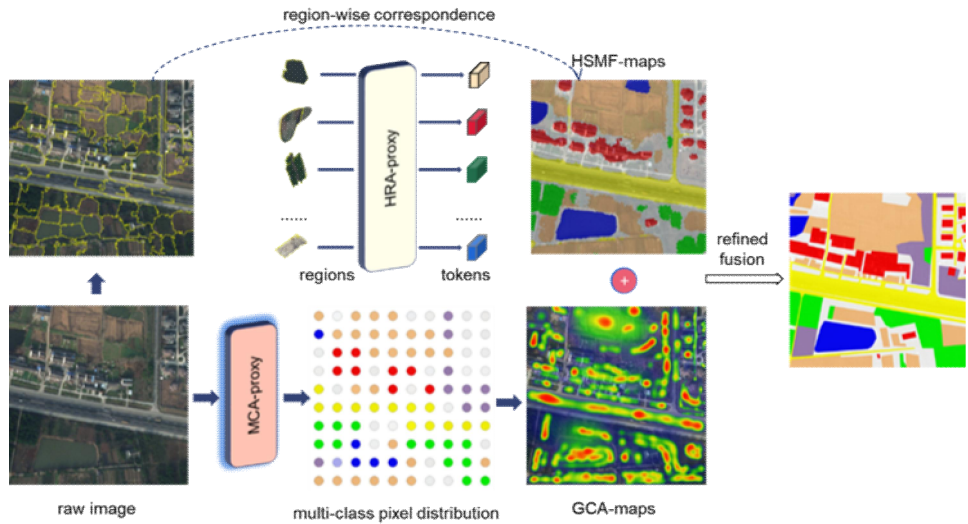

**Fig. 2.** Illustration of EMRA-proxy. We utilize HRA-proxy and MCA-proxy to respectively extract homogeneous semantic mask feature maps (HSMF-maps) and global class attention maps (GCA-maps). Finally, merge the two maps to achieve precise image segmentation.

**ViT-based Semantic Segmentation**. In recent years, Transformers have been effectively applied to a variety of computer vision tasks, including semantic segmentation. Utilizing attention mechanisms inspired by Transformers has enhanced segmentation model performance [31,32]. ViT [33] has garnered significant interest due to its efficacy in image classification and other computer vision tasks. Several enhanced ViT models with varying architectures, such as CaiT [34], and Swin Transformer [35], has been proposed. These models use hierarchical models to generate 2D features, recent studies have investigated the use of Transformer-based models for semantic segmentation tasks. For example, SETR [36] integrated ViT backbones into the semantic segmentation task, while Segmenter [37] employed Transformers to predict per-class masks. Other methods, such as SegFormer [38] and DPT [39], proposed hierarchical ViT backbones for dense prediction tasks. MCTformer [40] proposed employing a Multi-class Token Transformer to enhance weakly supervised semantic segmentation.

In subsequent research, researchers have actively explored the application of Transformers to address various remote-sensing tasks. Chen et al. [41] designed a dual-temporal image Transformer to model context in the spatiotemporal domain and improve the efficiency of change detection. Hong et al. [42] utilized Transformers for pan-sharpening of HRRS images, achieving optimal results both visually and quantitatively. Tao et al. [43] proposed an Enhanced Multi-scale Representation Transformer (EMRT), leveraging the advantages of convolutional operations and Transformers to enhance multi-scale representation learning.

CNN-based methods struggle to capture global information in remote sensing image segmentation, resulting in relatively imprecise segmentation outcomes, whereas ViT-based methods require more remote sensing image categories localization and boundaries. To overcome these limitations, we present a novel strategy. Specifically, we propose an EMRA-proxy to extract homogeneous semantic information of image regions and object category information for segmenting HRRS images.

## 3        Methods

Our proposed EMRA-proxy model aims to utilize homogeneous semantic information of image regions and object category information for segmenting HRRS images. An overview of the model overall architecture is presented in **Fig. 3**. Next, we will provide a comprehensive and detailed explanation of the EMRA proxy method.

In Section 3.1, we first introduce the overall architecture of the EMRA-proxy and the process of utilizing the Transformer to extract contextual information from regions. In Section 3.2, we provide a detailed explanation of the principle and process of HRA-proxy for extracting homogeneous region geometric features. In Section 3.3, we will describe the process of MCA-proxy extracting image category information, as well as delineate the process of generating the final segmented image.

### 3.1 Overall Network Architecture

HRRS images have complex background elements and ground objects of various shapes, so extracting local region features and category features is crucial to further improve segmentation performance. We proposed a method called EMRA-proxy to address the aforementioned issues. **Fig. 3** illustrates the overall structure of EMRA-proxy, where features are first extracted from the image using Transformer. These features are then separately utilized by MCA-proxy and HRA-proxy to obtain the image's category and region geometry information, respectively. Finally, the information obtained from MCA-proxy and HRA-proxy is merged to generate the final segmentation map.

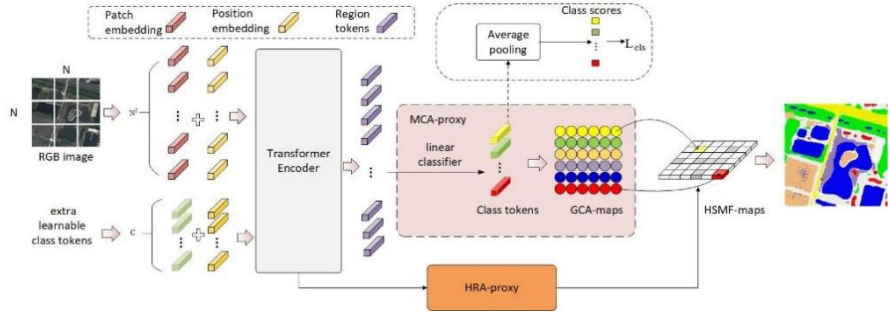

**Fig. 3.** Overview of our EMRA-proxy approach. We first split and transform an input RGB image into a sequence of patch tokens. Additionally, we propose to incorporate the learning of C extra class tokens, where C represents the number of classes. The C class tokens are concatenated with patch tokens and augmented with position embeddings (PE) which then undergo consecutive L transformer encoding layers. The HR proxy utilizes the outputs of the encoder's initial layers to generate a homogeneous semantic mask feature map (HSMF-map). Additionally, the MCA-proxy can generate a global class attention map (GCA-map) from the encoder's outputs. Finally, the fusion of HSMF-maps and GCA-maps produces the final segmentation results.

Transformer is a type of sequence-to-sequence model, that applies multi-layer self-attention on its computation primitives, i.e. tokens. In this work, we take full advantage of the Transformer architecture to learn the global context for images by directly modeling inter-region relations using self-attention. An image is split into a series of segments $N \times N$ patches, which are then transformed into a sequence of patch tokens $T_p \in R^{E \times D}$, where D is the embedding dimension, $E = N^2$. We propose multi-class tokens $T_{cls} \in R^{C \times D}$ to learn category features, where $C$ is the number of classes. The class embeddings $T_{cls}$ are randomly initialized and designated to a specific semantic class to generate the multi-class tokens. To learn positional information, position embedding $pos = [pos_1, ..., pos_N] \in R^{N \times D}$ is added to the sequence of patches $T_p$ and $T_{cls}$. The final input sequence of tokens is $T_{in} \in R^{(E+C) \times D}$.

The token sequence $T_{in}$ is processed by a transformer encoder [44] with L layers to generate contextualized encodings $T_L \in R^{M \times D}$. As shown in **Fig. 4**, each transformer layer consists of a multi-headed self-attention (MSA) block and a point-wise MLP block with two layers. Before and after each block, Layer norm (LN) is applied, and residual connections are added after each block. The formula of MSA and MLP formulas are as follows:

$$a_{\{l-1\}} = MSA\left(LN\left(T_{\{l-1\}}\right)\right) + T_{\{l-1\}} \tag{1}$$

$$T_l = MLP\left(LN\left(a_{\{l-1\}}\right)\right) + a_{\{l-1\}} \tag{2}$$

where $l \in \{1, ..., L\}$.

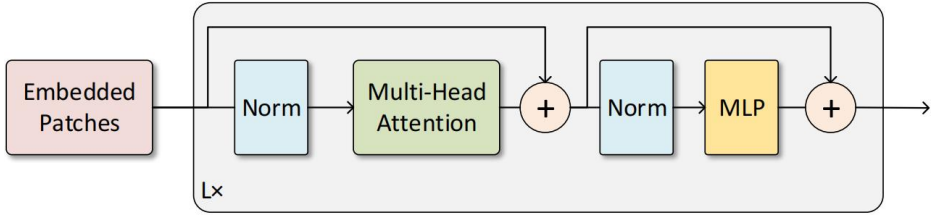

**Fig. 4.** Transformer encoder

Following the standard definition in [33,45] we experimented with four ViT model sizes (ViT-{Ti/16, S/16, B/16, L/16}) as backbones. The multi-class transformer encoder maps input patches with position encoding ($T_{in}$) to a contextualized encoding sequence ($T_l$) that contains semantic information. In the subsequent section, we will provide a detailed exposition of how MCA-proxy and HRA-proxy leverage the features extracted by the encoder.

### 3.2    Homogeneous regions attention proxy

Learning and describing the geometric features of regions is challenging due to their irregular shapes and varying scales. While predicting a binary mask for each region seems intuitive, it's not feasible for sequence-to-sequence models. Predicting full-size masks for each token would be computationally expensive. On the other hand, predicting small-size masks for local regions may compromise the coherence of the areas, resulting in overlaps or loss of information.

**Extracting geometric region.**

Revisiting classical superpixel segmentation [46], it is a process that entails grouping pixels based on low-level information, such as color, to form perceptually similar regions. This technique provides a foundational image representation for a variety of high-level visual tasks, including semantic segmentation. Following the findings outlined in [47], the semantic meaning of pixels with similar low-level attributes within close proximity should exhibit homogeneity. Leveraging this principle, we attempt to batch the pixel labeling by classifying superpixels.

To address this, we introduce a novel mechanism homogeneous regions attention proxy (HRA-proxy) for describing the geometric shapes of regions through pixel-to-token associations. HRA-proxy is to use region proxies to describe the geometric shapes of images with homogeneous superpixel groups. We start with an initial $H_g \times W_g$ grid, where $H_g \times W_g = N_g$. Each token resides on a grid cell. The cells themselves act merely as token position indicators, unrelated to the actual geometric shapes of the regions. We establish associations between pixels $p = (u, v)$ and tokens for region s by assigning them probabilities $q_s(p)$. We only associate a pixel p with tokens located in its neighborhood $N_p$, which satisfies:

$$\sum_{s \in N_p} q_{s(p)} = 1. \qquad (3)$$

**Fig. 5** illustrates this concept, where each pixel p is allocated to one of nine regions. Mathematically, we describe the inter-pixel associations using an affinity graph $G \in \mathbb{R}^{(H_g \times h) \times (W_g \times w) \times |N_p|}$ that is agnostic to categories, portraying the actual geometric shapes of the regions. The dimensions $(H_g \times h, W_g \times w)$ represent the size of the output segmentation map, where $(h, w)$ denotes the relative stride of the initial token grid. Observations indicate that a neighborhood size of $3 \times 3$ and $|N_p| = 9$ is effective for all model sizes. Based on empirical observations [47], we set the domain size to 3x3 and Np=9, which is suitable for images of all model sizes.

For a more intuitive understanding, the "core" of a region is represented by its location on the initial $H_g \times W_g$ cell, whereas the surrounding pixels are represented by probabilities. Following the constraint of equation (3), the probabilistic regions form a tessellation that completely encompasses the image plane without any gaps or overlaps.

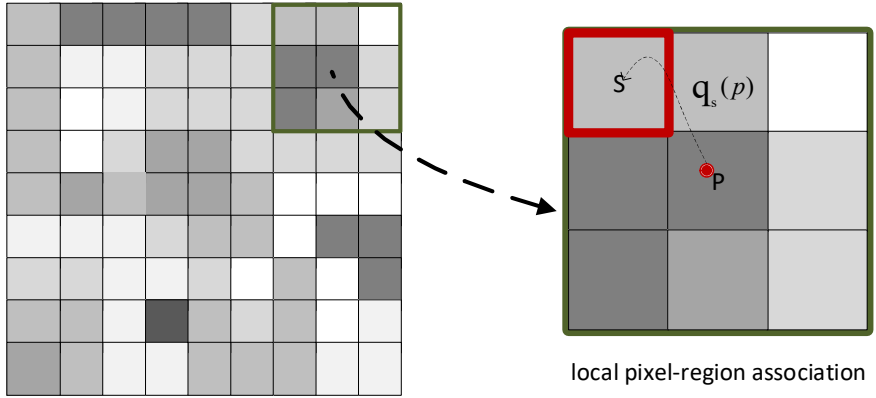

local pixel-region association

**Fig. 5.** Describing region geometrics by local pixel-region association.

### HRA-proxy Implementation.

As illustrated in **Fig. 6**, We utilize HRA-proxy to learn geometric descriptions of G, using embed region features as input. To embed region features into tokens, we employ a portion of the ViT backbone as token heads, generating E tokens with dimension D. Through comparative experiments across different layers, we determine the optimal M, selecting the first M Transformer layers of ViT as token heads.

In detail, the convolution module comprises one depth-wise convolution layer and one convolution layer. The depth-wise convolution layer with a kernel size of 3 × 3, is then followed by a 1 × 1 convolution layer. After convolution, the HSMF-maps are subjected to Softmax activation, which generates normalized probabilities. Convolutional operations excel at capturing local information within images and extracting geometric features of regions.

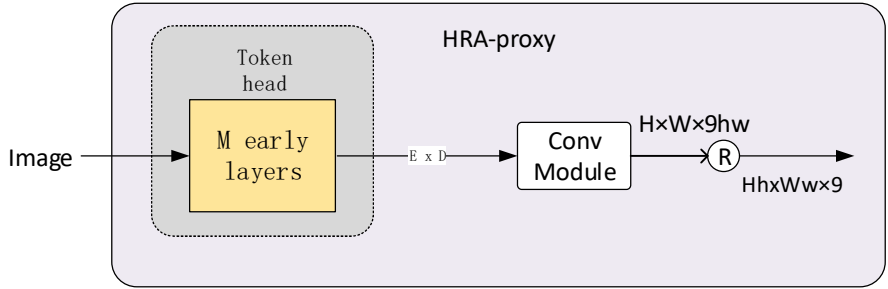

**Fig. 6.**    lustration of the HRA-proxy, where Conv Module stands for depthwise separable convolution and R stands for reshape and rearrange of dimensions.

### 3.3    Multi-class Attention proxy

During training, to ensure that tokens from different classes can learn distinct class-specific features, we incorporate additional class tokens into the input, as depicted in **Fig. 3**. In the learning process of the Transformer encoder, we directly compute the classification loss to establish a strong connection between each token and its class label. By utilizing a Multi-class Attention proxy (MCA-proxy), we extract class scores for different pixels to serve as proxies for the class of region pixels, thereby guiding image segmentation.

As illustrated in **Fig. 7**, MCA-proxy uses the standard self-attention layer to capture the long-range dependencies between tokens. The self-attention mechanism consists of three point-wise linear layers that map tokens to intermediate representations, including queries $Q \in \mathbb{R}^{(E+C) \times d}$, keys $K \in \mathbb{R}^{(E+C) \times d}$ and values $V \in \mathbb{R}^{(E+C) \times d}$. We utilize the Scaled Dot-Product Attention [44] mechanism to calculate the attention scores between queries and keys. Each resultant token is derived from a weighted sum of all tokens, with attention scores serving as the weights. This can be formulated as:

$$Attention(Q, K, V) = softmax\left(\frac{QK^T}{\sqrt{D}}\right)V, \qquad (4)$$

where we can obtain a global class attention map (GCA-map) $A_{t2t} \in \mathbb{R}^{(C+M)\times(C+M)}$ and $A_{t2t} = softmax\left(\frac{QK^T}{\sqrt{d}}\right)$. We aggregate the transformer attentions from the last P layers and multiple heads to produce a final attention map. From this map, we derive class-specific object localization maps and a patch-level pairwise affinity map using the class-to-patch and patch-to-patch attentions, respectively. The patch-level pairwise affinity can refine the GCA-map, resulting in enhanced image segmentation. We conducted a series of comparative experiments to determine how to choose K, as illustrated in **.** The Impact of Parameter P on the Accuracy of MCA-proxy..

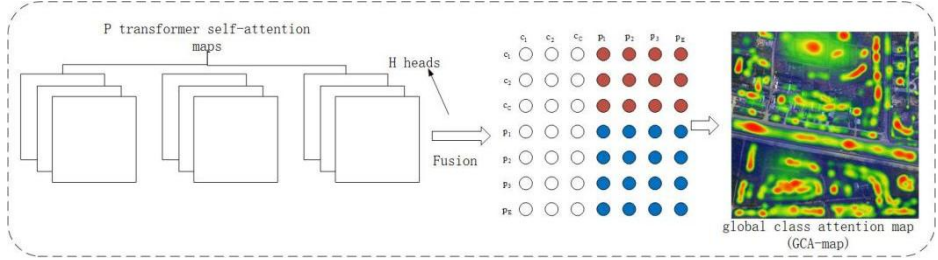

**Fig. 7.** We aggregate the transformer attentions from the last P layers and multiple heads to produce a final attention map.

Finally, we fuse the GCA-map obtained from the MCA-proxy with the HSMF-map generated by the HRA-proxy to produce the final segmentation results, as shown in **Fig. 3**. The segmentation results are efficiently drawn on a per-pixel basis. The class mask for a given pixel p = (u, v) is calculated directly:

$$M'[u, v] = \sum_{s \in N_p} A_{t2t}(s) \cdot q_s(p) \qquad (5)$$

where $M' \in R^{(H_{gh})\times(W_{gw})\times C}$ is the output logits map, $A_{t2t}(s)$ corresponds to the class logits for region s obtained from the multi-class token. We set $(w, h) = (4,4)$ to generate the ×4 stride logits map for Ours-x/16 models. Our model is trained end-to-end using cross-entropy loss without class balance or hard example mining techniques. To get the final segmentation map, we first enhance the resolution of the intermediate feature map $M'$ using bilinear interpolation. We then apply Softmax and layer normalization to obtain per-pixel class scores, which are used to generate the final segmentation map.

## 4    Experiments and results

### 4.1    Datasets and Metrics

To evaluate the performance of our proposed method, we conducted experiments on three open datasets, including two high-resolution datasets Potsdam and Vaihingen, as well as the more challenging LoveDA dataset. We used three evaluation metrics including mIoU, OA, and F1 score. Below is a detailed description of the datasets and evaluation metrics.

**LoveDA [5]:** The dataset is a comprehensive collection of high-resolution (HSR) images with meticulous annotations of land-cover objects from three Chinese cities: Nanjing, Changzhou, and Wuhan. The dataset consists of 5987 images, of which 2713 are urban scenes, and 3274 are rural scenes, offering a diverse range of environments for computer vision research. The dataset is divided into training, validation, and testing sets, with 2522, 1669, and 1796 images, respectively. The dataset is annotated with 166768 objects across seven semantic categories, as shown in **Table 1**.

**Table 1.** RGB values of the categories.

| category | RGB Value |
| --- | --- |
| background | (255, 255, 255) |
| building | (255, 0, 0) |
| road | (255, 255, 0) |
| water | (0, 0, 255) |
| barren | (159, 129, 183) |
| forest | (0, 255, 0) |
| agriculture | (255, 195, 128) |

**Postdam:** The Potsdam dataset consists of 38 high-resolution images collected from urban scenes, each with dimensions of $6000 \times 6000$ pixels and a spatial pixel resolution of 5 centimeters. The source images include large building blocks, narrow streets, and dense settlement structures. Ground objects are classified into six categories: impervious surfaces, buildings, low vegetation, trees, cars, and clutter/background. We used 24 images for training, and the remaining 14 images were used for testing.

**Vaihingen:** The Vaihingen dataset showcases a small village with many independent multi-story buildings and small multi-story buildings. The dataset consists of 33 orthoimage blocks, each with a spatial resolution of 9 centimeters and an average size of $2494 \times 2064$ pixels. These images are densely classified into six categories, the same as those in the Potsdam dataset. In experiments, 16 images are used for training, and the remaining 17 images are used for testing.

**Metrics.** To comprehensively measure the performance of our proposed model, the mean intersection over union (mIoU), the overall accuracy (OA), and the F1 score are used as evaluation metrics. Among them, mIoU denotes the ratio of intersection and union of the true and predicted values, OA refers to the ratio of correctly predicted pixels by category to the total pixels, and the F1 score is the harmonic mean between the accuracy and recall of the model.

$$mIoU = \frac{1}{N}\sum_{k=1}^{N}\frac{TP_k}{TP_k+FP_k+FN_k} \qquad (6)$$

$$OA = \frac{\sum_{k=1}^{N} TP_k}{TP_k + FP_k + TN_k + FN_k} \tag{7}$$

$$F1 = 2 \times \frac{Precision \times Recall}{Precision + Recall} \tag{8}$$

In addition, precision and recall are defined as:

$$Precision = \frac{1}{N} \sum_{k=1}^{N} \frac{TP_k}{TP_k + FP_k} \tag{9}$$

$$Recall = \frac{1}{N} \sum_{k=1}^{N} \frac{TP_k}{TP_k + FN_k} \tag{10}$$

where $TP_k, FP_k, TN_k, and\ FN_k$ represent the counts of true positive, false positive, true negative, and false negative pixels for the object categorized as class k, respectively.

### 4.2    Implementation Details

ViT backbone. We employ the Vision Transformer (ViT) as our encoder, which is available in four various sizes: Large, Base, Small, and Tiny, as shown in **Table 2**. The parameter count of the transformer encoder varies with different layers and token sizes. Specifically, we fix the head size to 64 in the multi-headed attention mechanism, while the number of heads is determined by the token size divided by the head size. Additionally, the hidden size of the MLP following MSA is set to four times the token size. We use weights pre-trained on ImageNet21k [48] following recent works [37,47].

**Table 2.** Details of Transformer variants.

| method | backbone | Layers | Token size | Heads | Params |
|--------|----------|--------|-----------|-------|--------|
| EMRA-proxy-T | ViT-Ti/16 | 12 | 192 | 3 | 6.69M |
| EMRA-proxy-S | ViT-S/16 | 12 | 384 | 6 | 26.19M |
| EMRA-proxy-B | ViT-B/16 | 12 | 768 | 12 | 103.1M |
| EMRA-proxy-L | ViT-L/16 | 24 | 1024 | 16 | 334.3M |

**Train Details.** Our method is implemented using the publicly available mmsegmentation [49] codebase, with minimal modifications made to its default settings that are widely adopted by the community. We use input sizes of 512 × 512 for LoveDA, Potsdam and Vaihingen. We train our "Large" model using a 640 × 640 crop following [37,38,47]. We use a batch size of 16 and train for 100 epochs. We use stochastic gradient descent (SGD) [50] as the optimizer with a base learning rate $\gamma_0$ and set weight decay to 0. Inspired by the seminal work of segmenter [37], we employ the "poly" learning rate decay, which is defined as $\gamma = \gamma_0 \left(1 - \frac{N_{epoch}}{N_{total}}\right)^{0.9}$, where $N_{epoch}$ and $N_{total}$ represent the current epoch number and the total iteration number, respectively. Our work sets the base learning rate $\gamma_0$ to $10^{-3}$. We maintain the default settings for data augmentations and all other training parameters in [49]. To fine-tune the pre-trained models for semantic segmentation, we employ the standard pixel-wise cross-entropy loss without weight rebalancing.

**Inference**. In order to accommodate images of varying sizes, we employ a sliding window technique that is congruent with the training dimensions. For multi-scale inference, we adhere to established standard practice [51] by utilizing rescaled iterations of the images with scaling factors of (0.5, 0.75, 1.0, 1.25, 1.5, 1.75), alongside left-right mirroring. Subsequently, the outcomes of these rescaled iterations are averaged to derive the ultimate prediction.

### 4.3     Results

**Ablation study.**

In this section, we ablate various variants of our methodology on the LoveDA validation set. Following recent works [37,47], we augment bare ViT by adding a linear classifier to generate per-patch predictions. Following the standard segmentation pipeline, these predictions are upsampled to the image size for training and inference.

*Ablation Study for Backbone.*
In our study, we conducted an extensive ablation analysis to investigate the impact of different backbone architectures on the performance of our method. The results of this analysis, summarized in **Table 3**, clearly illustrate the significance of backbone capacity in determining the overall performance of the model. Specifically, we observed that increasing the capacity of the backbone consistently leads to improved performance across various evaluation metrics.

To balance performance with computational efficiency, we also evaluated the trade-off between mIoU and Floating Point Operations (FLOPs). As depicted in **Fig. 8**, our analysis indicates that the EMRA-proxy-B configuration emerges as the optimal choice. This configuration strikes an optimal balance between achieving high mIoU scores and minimizing computational complexity, making it well-suited for practical deployment in real-world scenarios.

**Table 3.** Performance comparison of our models with varying backbones and input patch sizes on LoveDA validation set.

| method | backbone | FLOPs | aAcc(%) | mIoU(SS/MS %) | |
|---|---|---|---|---|---|
| EMRA-proxy-T | ViT-Ti/16 | 4.47G | 70.79 | 52.4 | 53.59 |
| EMRA-proxy-S | ViT-S/16 | 17.78G | 71.29 | 52.87 | 54.62 |
| EMRA-proxy-B | ViT-B/16 | 70.36G | 72.13 | 54.50 | 55.22 |
| EMRA-proxy-L | ViT-L/16 | 356.01G | 72.11 | 54.60 | 55.47 |

*Ablation Study for Improvement Strategies.*

proxy on the segmentation performance using ViT-base as the benchmark. As shown in **Table 4**. The results indicate a significant improvement in mIoU compared to the baseline. The MCA-proxy and HRA-proxy contributed to varying degrees of improvement over the baseline. The optimal performance was achieved by combining the two, with single-scale (SS) mIoU and multi-scale (MS) mIoU reaching 54.50% and 55.22% respectively. **Fig. 9** shows the efficacy curves of how the HRA-proxy and MCA-proxy strategies, based on the ViT-B backbone, work during the training phase. We can observe that combining the two proxy modules results in varying degrees of improvement across different categories, thereby achieving excellent segmentation performance.

Upon comparing the F1 scores, precision, and recall of various categories as illustrated in **Fig. 10**, it is evident that most categories demonstrate strong performance in precision and recall. However, in the Barren category, the precision is high while the recall is low, indicating the presence of missed detections. This is mainly because the data volume of the Barren category is relatively small, resulting in weaker classification capability.

We present a qualitative comparison of the ablation study in **Fig. 11**. The figure illustrates the enhanced semantic segmentation performance achieved by the MCA-proxy and HRA-proxy methods across various scenarios. Through analysis of the depicted black boxes, it is evident that the MCA-proxy effectively localizes objects from diverse categories, resulting in improved segmentation outcomes for geometrically regular objects, indicative of its sensitivity to object categories. However, when faced with objects exhibiting complex geometric shapes, the MCA-proxy struggles to accurately segment fine details in local regions. To mitigate this limitation, we leverage the HRA-proxy to extract homogeneous semantic mask feature maps. As highlighted in the white box within the figure, the HRA-proxy facilitates finer segmentation of objects, particularly in regions with intricate geometry. Ultimately, we integrate both methods (EMRA-proxy) to achieve a more accurate segmentation effect, capitalizing on the strengths of each proxy module. This comprehensive approach can achieve strong segmentation performance in various object categories and geometries, validating the effectiveness of our proposed method.

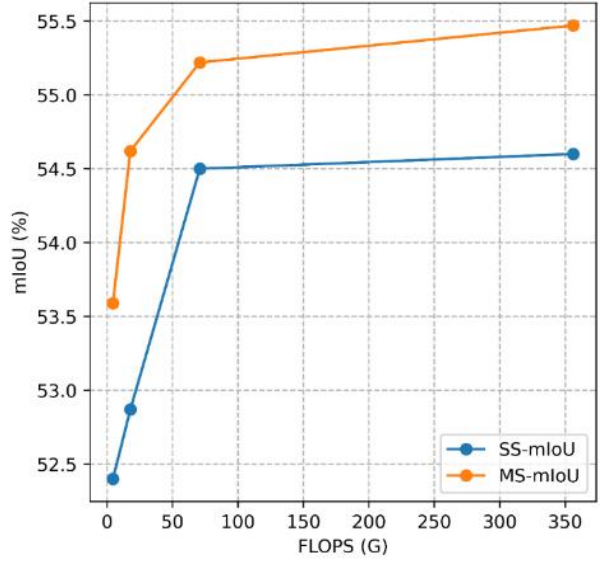

**Fig. 8.** Performance vs. GFLOPs using Ours approach on LoveDA val split. We report results with single-scale (SS) and multi-scale (MS) inference.

**Table 4.** Impact of MCA-proxy and HRA-proxy on Segmentation Performance. We report mIoU results of EMRA-proxy-B/16 on LoveDA.

| method | MCA-proxy | HRA-proxy | IoU per category (%) | | | | | | | mIoU (%) |
|---|---|---|---|---|---|---|---|---|---|---|
| | | | Background | Building | Road | Water | Barren | Forest | Agriculture | |
| ViT-B/16 | - | - | 42.93 | 44.97 | 36.85 | 51.85 | 14.8 | 33.01 | 42.63 | 38.15 |
| EMRA-proxy-B/16 | + | - | 51.43 | 60.4 | 55.81 | 69.44 | 30.86 | 40.66 | 59.58 | 52.6 |
| EMRA-proxy-B/16 | - | + | 53.01 | 60.49 | 56.19 | 70.28 | 27.99 | 41.61 | 60.1 | 52.81 |
| EMRA-proxy-B/16 | + | + | 54.54 | 65.14 | 57.56 | 71.45 | 29.04 | 45.66 | 63.11 | 55.22 |

*Depth of HRA-proxy Token Head.*

As shown in **Fig. 6**, we utilize the first M layers of the Transformer as the Token head. Through experimentation, as shown in **Table 5**, we observed that the performance is optimal within the initial layers, particularly within layers 3-5. Deeper or shallower configurations negatively affect performance. According to the results, it can be seen that if the M layer is too deep, the information loss will increase, and it is difficult for HRA-proxy to obtain accurate regional geometric information. If M is too shallow, the HRA-proxy module has limited capabilities and cannot extract the required features. Therefore, after experiments, it was verified that when M=3, HRA-proxy can obtain regional feature information to the greatest extent.

*The Impact of Parameter P on the Accuracy of MCA-proxy.*

The segmentation accuracy of the global category attention map generated by fusing the last P layers of Transformer layers was evaluated. From **Fig. 12**, it can be observed that the segmentation accuracy is highest at P=4. Excessive aggregation of layers leads the model to learn numerous features unrelated to the categories, thereby diminishing the accuracy of class-specific feature extraction. Conversely, insufficient layer aggregation results in the loss of significant information, making it challenging to learn precise category information.

**Comparison with state-of-the-art.**

In this section, we compare the performance of EMRA-proxy with respect to the state-of-the-art methods on LoveDA, Potsdam and Vaihingen datasets.

*Comparison on the LoveDA Dataset.*

We compared the segmentation performance of EMRA-proxy with some of the latest methods on the LoveDA dataset. These methods include both CNN-based approaches and Transformer-based methods. They followed their own optimal parameters and loss function settings in the comparison experiments. Unlike typical datasets, LoveDA has a larger data volume and more complex backgrounds. The overall results for each method are presented in **Table 6.** Comparison to state-of-the-art methods on LoveDA test set.. Our method achieved an mIoU of 55.22% an OA of 73.29% and an mF1 of 70.14% representing a 2.08% improvement in mIoU compared to the state-of-the-art methods.

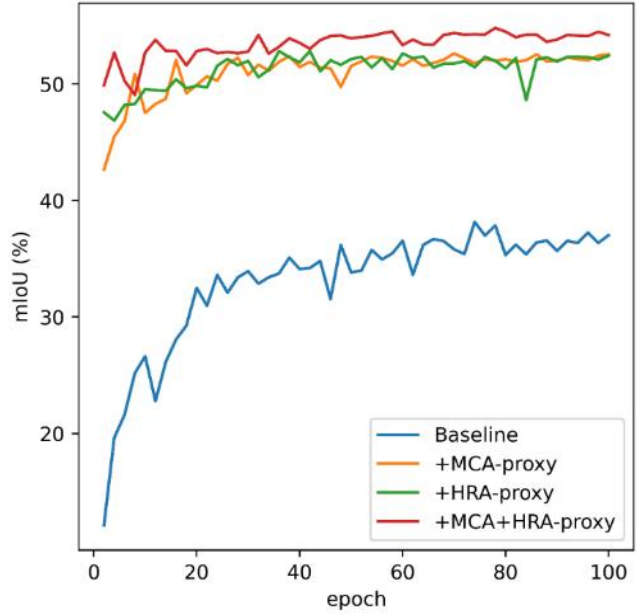

**Fig. 9.** The curves demonstrate the effectiveness of MCA-proxy and HRA-proxy, recording the mIoU value with epoch during the training phase of different module combinations in EMRA-proxy.

**Table 5.** Selecting the Depth of HRA-proxy Token Head for Region Learning. We report single scale mIoU results of EMCR-proxy on LoveDA

| M | 0 | 3 | 5 | 7 | 9 | 12 |
|---|---|---|---|---|---|---|
| mIoU(%) | 53.45 | 55.22 | 54.52 | 53.4 | 52.59 | 51.20 |

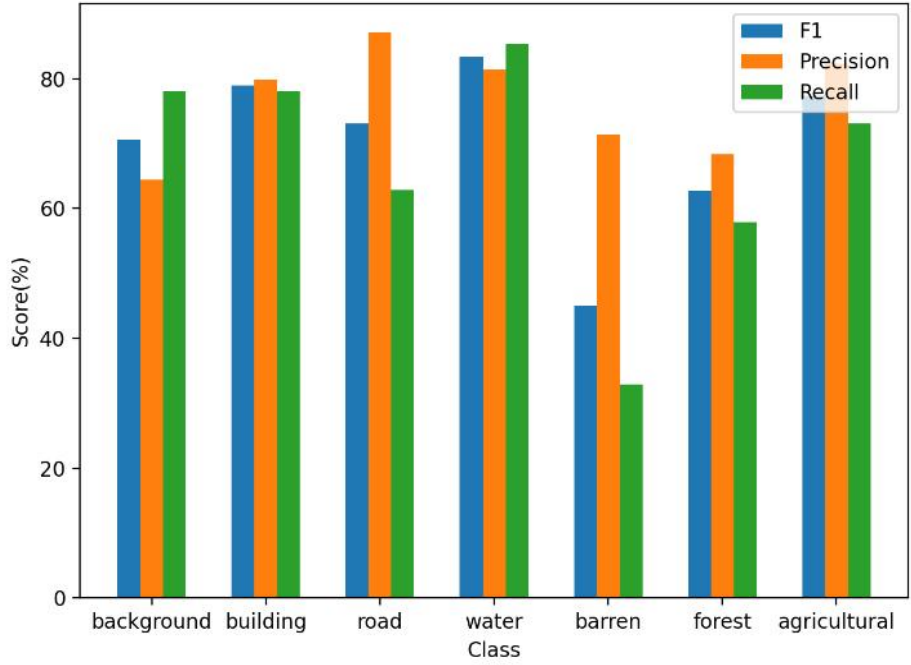

**Fig. 10.** The distribution of F1 scores, precision, and recall across different classes.

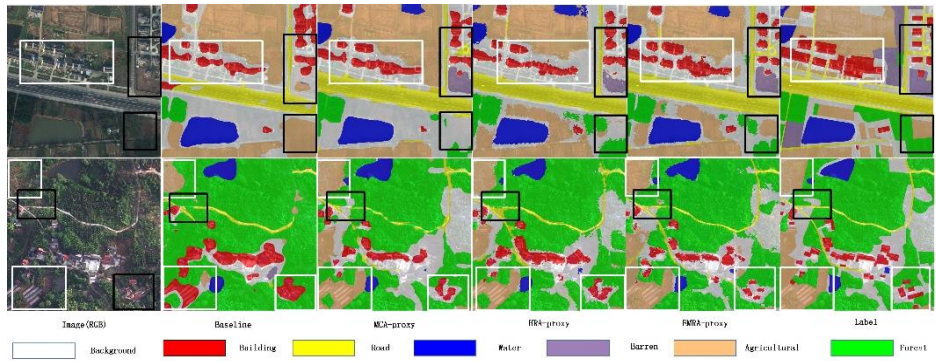

**Fig. 11.** Qualitative comparison of ablation study on LoveDA. We compare the influence of different network architectures on remote sensing image segmentation. Best viewed zoom in.

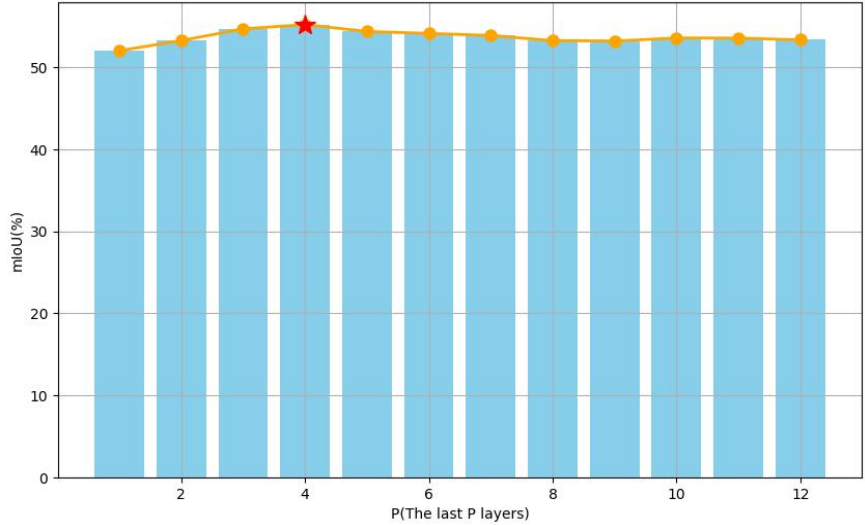

**Fig. 12.** The Impact of Parameter P on the Accuracy of MCA-proxy.

Furthermore, our proposed method achieved the highest IoU and F1 values for most categories, particularly for categories with significant intra-class variations, such as water, buildings, and forests. This validates the effectiveness of MCA-proxy for extracting multi-class category information and HRA-proxy for extracting complex geometric feature information.

**Fig. 13** provides some qualitative comparison examples. As observed, some methods perform well in rural scenes but poorly in urban scenes, such as Reproxy. In contrast, HRNet and EMRT perform well only in urban scenes. The proposed EMRA-proxy achieves the best results in both scenarios, as demonstrated by more accurate predictions for large-scale objects (e.g., agriculture and forests) and small-scale multi-scale objects (e.g., roads and building) without the need for any post-processing to maintain accuracy along the boundaries. These results indicate that EMRA-proxy is more accurate in segmenting edges between neighboring objects and preserves spatial details effectively.

*Comparison on the Potsdam Datasets.*

To verify the universality of the EMRA-proxy method, we conducted experiments on the ISPRS Potsdam dataset, and the results are shown in **Table 7.** Comparison to state-of-the-art methods on Potsdam test set.. Across all six categories, the EMRA-proxy method demonstrates superior performance. Due to the high visual similarity between low vegetation and trees, which often appear in adjacent areas, many methods tend to misclassify low vegetation areas as trees. However, EMRA-proxy is able to accurately distinguish between them. This is attributed to the effectiveness of EMRA-proxy's HRA-proxy and MCA-proxy, which excel at distinguishing complex categories and geometric regions in images. EMRA-proxy combines CNN and Transformer architectures to some extent, leveraging the strengths of both architectures. It achieves excellent performance in segmenting both small and large targets, particularly excelling in tasks such as segmenting impervious surface, building, tree, and car.

We also compared these state-of-the-art methods on the ISPRS Vaihingen dataset. This dataset contains relatively fewer training samples, smaller image resolutions, and more small-scale objects, leading to inferior performance of most compared methods compared to the Potsdam dataset. The results recorded in **Table 8** show that EMRA-proxy has reached a new level, achieving an mIoU of 70.03% an OA of 87.14% and an mF1 of 81.96%. This demonstrates the generality of our approach, which is effective for remote sensing image segmentation in various scenarios.

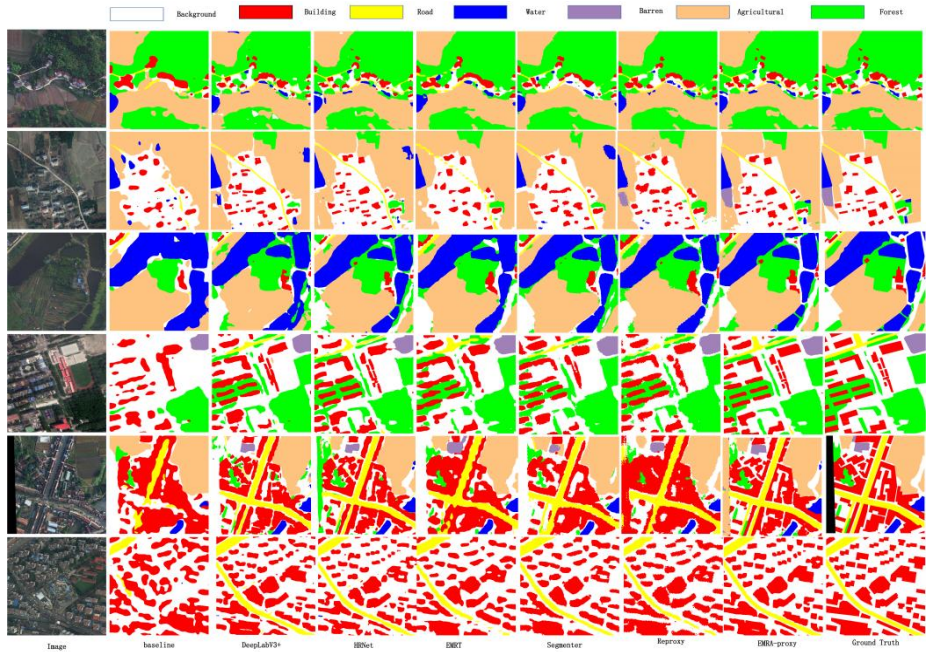

**Fig. 13.** Qualitative comparison with different methods on the LoveDA dataset. The top three rows are selected from the rural scene, and the bottom three rows are selected from the urban scene. Best viewed zoom in.

**Table 6.** Comparison to state-of-the-art methods on LoveDA test set.

| method | Publication | IoU per category (%) | | | | | | | mIoU (%) | OA (%) | mF1(%) |
|---|---|---|---|---|---|---|---|---|---|---|---|
| | | Backg round | Buildi ng | Road | Water | Barren | Forest | Agricult ure | | | |
| FCN8S [13] | CVPR15 | 42.60 | 49.51 | 48.05 | 73.09 | 11.84 | 43.49 | 58.30 | 46.69 | 66.07 | 62.63 |
| DeepLabV3+ [51] | ECCV18 | 42.97 | 50.88 | 52.02 | 74.36 | 10.40 | 44.21 | 58.53 | 47.62 | 68.10 | 64.60 |

| method | Publication | | | | | | | | mIoU (%) | OA (%) | mF1 (%) |
|---|---|---|---|---|---|---|---|---|---|---|---|
| UNet [14] | MICCAI15 | 43.06 | 52.74 | 52.78 | 73.08 | 10.33 | 43.05 | 59.87 | 47.84 | 66.50 | 64.70 |
| UNet++ [52] | DLMIA18 | 42.85 | 52.58 | 52.82 | 74.51 | 11.42 | 44.42 | 58.80 | 48.20 | 67.58 | 65.25 |
| PSPNet [53] | CVPR17 | 44.40 | 52.13 | 53.52 | 76.50 | 9.73 | 44.07 | 57.85 | 48.31 | 67.85 | 65.20 |
| HRNet [54] | CVPR19 | 44.61 | 55.34 | 57.42 | 73.96 | 11.07 | 45.25 | 60.88 | 49.79 | 67.80 | 65.50 |
| Segmenter [37] | ICCV21 | 51.43 | 60.4 | 55.81 | 69.44 | 30.86 | 40.66 | 59.58 | 52.60 | 68.94 | 66.10 |
| RegProxy [47] | CVPR22 | 50.96 | 60.24 | 56.62 | 70.79 | 28.63 | 42.31 | 59.96 | 52.79 | 70.23 | 68.54 |
| TrSeg [55] | PRL21 | 50.08 | 59.23 | 48.83 | 61.52 | 28.67 | 39.48 | 49.60 | 48.20 | 66.41 | 64.43 |
| $I^2HN$ [56] | TIP23 | 51.96 | 61.56 | 55.84 | 70.88 | 28.55 | 42.31 | 60.98 | 53.14 | 71.56 | 69.33 |
| EMRT [43] | TGRS23 | 50.58 | 60.48 | 52.78 | 66.73 | 32.88 | 40.28 | 52.51 | 50.89 | 68.27 | 66.78 |
| **Ours** | - | **54.54** | **65.14** | **57.56** | **71.54** | **29.04** | **45.66** | **63.11** | **55.22** | **73.29** | **70.14** |

**Table 7.** Comparison to state-of-the-art methods on Potsdam test set.

| method | Publication | IoU per category (%) | | | | | | mIoU (%) | OA (%) | mF1 (%) |
|---|---|---|---|---|---|---|---|---|---|---|
| | | Impervious Surface | Building | Low Vegetation | Tree | Car | Background | | | |
| FCN8S [13] | CVPR15 | 81.64 | 89.11 | 71.36 | 73.34 | 79.32 | 33.87 | 71.44 | 87.17 | 81.85 |
| PSPNet [53] | CVPR17 | 82.68 | 90.17 | 72.72 | 74.00 | 80.56 | 35.86 | 72.67 | 87.90 | 82.75 |
| DeepLabV3+ [51] | ECCV18 | 82.10 | 89.16 | 71.94 | 73.65 | 79.97 | 36.70 | 72.25 | 87.44 | 82.59 |
| TrSeg [55] | PRL21 | 82.39 | 90.15 | 72.37 | 74.20 | 80.58 | 36.21 | 72.65 | 87.82 | 82.79 |
| RegProxy [47] | CVPR22 | 83.10 | 90.50 | 72.50 | 73.87 | 81.76 | 38.31 | 73.34 | 88.15 | 82.78 |
| EMRT [43] | TGRS23 | 83.27 | 90.22 | 72.59 | 74.26 | 82.34 | 39.04 | 73.62 | 88.12 | 83.59 |
| **Ours** | - | **83.89** | **91.23** | **72.75** | **75.10** | **83.23** | **40.16** | **74.39** | **88.87** | **83.96** |

**Table 8.** Comparison to state-of-the-art methods on Vaihingen test set.

| method | Publication | IoU per category (%) | | | | | | mIoU (%) | OA (%) | mF1 (%) |
|---|---|---|---|---|---|---|---|---|---|---|
| | | Impervious Surface | Building | Low Vegetation | Tree | Car | Background | | | |
| FCN8S [13] | CVPR15 | 78.11 | 84.82 | 63.78 | 75.08 | 53.38 | 38.05 | 65.54 | 85.86 | 77.98 |
| PSPNet [53] | CVPR17 | 79.16 | 85.90 | 64.36 | 74.94 | 60.93 | 40.84 | 67.69 | 86.26 | 79.75 |
| DeepLabV3+ [51] | ECCV18 | 78.62 | 88.07 | 64.47 | 75.43 | 58.69 | 40.41 | 67.28 | 86.27 | 79.41 |
| TrSeg [55] | PRL21 | 79.40 | 86.31 | 63.56 | 75.28 | 61.87 | 39.93 | 67.72 | 86.37 | 79.72 |
| RegProxy [47] | CVPR22 | 80.20 | 86.52 | 64.90 | 74.14 | 66.71 | 40.45 | 68.82 | 86.10 | 81.54 |
| EMRT [43] | TGRS23 | 80.44 | 86.35 | 64.87 | 76.28 | 66.84 | 43.98 | 69.79 | 86.97 | 81.38 |
| **Ours** | - | **80.53** | **87.47** | **64.97** | **76.80** | **67.24** | **43.16** | **70.03** | **87.14** | **81.96** |

# 5    Conclusion

In this paper, we propose a breakthrough semantic segmentation method EMRA-proxy for remote sensing images, specifically tailored for remote sensing land cover images, which is integrated with HRA-proxy and MCA-proxy. Our method first utilizes Vision Transformer to effectively capture contextual information and then exploits the introduction of HRA-proxy to allow the interpretation of images into adaptive segments with homogeneous semantics, improving segmentation accuracy for targets with complex geometric shapes. In addition, we use MCA-proxy to extract multi-category features in complex images to further improve the segmentation effect of remote sensing images. The experiments on three public datasets demonstrate that our method significantly outperforms existing CNN and Transformer-based methods in cases with comparable parameters, and shows robustness across various scenarios of remote sensing images.

# References

1. L. Lopez-Fuentes, C. Rossi, and H. Skinnemoen, "River segmentation for flood monitoring," in 2017 IEEE international conference on big data (Big Data). IEEE, 2017, pp. 3746–3749.
2. D. Su, H. Kong, Y. Qiao, and S. Sukkarieh, "Data augmentation for deep learning based semantic segmentation and crop-weed classification in agricultural robotics," Computers and Electronics in Agriculture, vol. 190, p. 106418, 2021.
3. C. Liu, S. Du, H. Lu, D. Li, and Z. Cao, "Multispectral semantic land cover segmentation from aerial imagery with deep encoder–decoder network," IEEE Geoscience and Remote Sensing Letters, vol. 19, pp. 1–5, 2020.
4. R. Cao, W. Tu, C. Yang, Q. Li, J. Liu, J. Zhu, Q. Zhang, Q. Li, and G. Qiu, "Deep learning-based remote and social sensing data fusion for urban region function recognition," ISPRS Journal of Photogrammetry and Remote Sensing, vol. 163, pp. 82–97, 2020.
5. J. Wang, Z. Zheng, A. Ma, X. Lu, and Y. Zhong, "Loveda: A remote sensing land-cover dataset for domain adaptive semantic segmentation," in Proceedings of the Neural Information Processing Systems Track on Datasets and Benchmarks 1, NeurIPS Datasets and Benchmarks 2021, December 2021, virtual, 2021.
6. J. Ali, M. Aldhaifallah, K. S. Nisar, A. A. Aljabr, and M. Tanveer, "Regularized least squares twin svm for multiclass classification," Big Data Research, vol. 27, p. 100295, 2022.
7. A. Radman, N. Zainal, and S. A. Suandi, "Automated segmentation of iris images acquired in an unconstrained environment using hog-svm and growcut," Digital Signal Processing, vol. 64, pp. 60–70, 2017.
8. S. Qi, J. Ma, J. Lin, Y. Li, and J. Tian, "Unsupervised ship detection based on saliency and s-hog descriptor from optical satellite images," IEEE geoscience and remote sensing letters, vol. 12, no. 7, pp. 1451–1455, 2015.
9. J. Chen, J. Zhu, Y. Guo, G. Sun, Y. Zhang, and M. Deng, "Unsupervised domain adaptation for semantic segmentation of high-resolution remote sensing imagery," IEEE Transactions on Geoscience and Remote Sensing, vol. 60, pp. 1–15, 2022.

10. X. Cheng, X. He, M. Qiao, P. Li, S. Hu, P. Chang, and Z. Tian, "Enhanced contextual representation with deep neural networks for land cover classification based on remote sensing images," International Journal of Applied Earth Observation and Geoinformation, vol. 107, p. 102706, 2022.

11. L. Zhang, Y. Wang, and Y. Huo, "Object detection in high-resolution remote sensing images based on a hard-example-mining network," IEEE Transactions on Geoscience and Remote Sensing, vol. 59, no. 10, pp. 8768–8780, 2020.

12. C. Zhang, Y. Feng, L. Hu, D. Tapete, L. Pan, Z. Liang, F. Cigna, and P. Yue, "A domain adaptation neural network for change detection with heterogeneous optical and SAR remote sensing images," International Journal of Applied Earth Observation and Geoinformation, vol. 109, p. 102769, 2022.

13. J. Long, E. Shelhamer, and T. Darrell, "Fully convolutional networks for semantic segmentation," pp. 3431–3440, 2015.

14. Ronneberger, O., Fischer, P., Brox, T.: U-net: Convolutional networks for biomedical image segmentation. In: Medical Image Computing and Computer-Assisted Intervention - MICCAI 2015 - 18th International Conference Munich, Germany, October 5 - 9, 2015, Proceedings, Part III, ser. Lecture Notes in Computer Science, vol. 9351, pp. 234–241. Springer, 2015.

15. R. Li, S. Zheng, C. Duan, J. Su, and C. Zhang, "Multistage attention ResU-Net for semantic segmentation of fine-resolution remote sensing images," IEEE Geoscience and Remote Sensing Letters, vol. 19, pp. 1–5, 2021.

16. C. Peng, X. Zhang, G. Yu, G. Luo, and J. Sun, "Large kernel matters–improve semantic segmentation by global convolutional network," in Proceedings of the IEEE conference on computer vision and pattern recognition, 2017, pp. 4353–4361.

17. L.-C. Chen, Y. Zhu, G. Papandreou, F. Schroff, and H. Adam, "Encoder-decoder with atrous separable convolution for semantic image segmentation," in Proceedings of the European conference on computer vision (ECCV), 2018, pp. 801–818.

18. H. Zhao, J. Shi, X. Qi, X. Wang, and J. Jia, "Pyramid scene parsing network," in Proceedings of the IEEE conference on computer vision and pattern recognition, 2017, pp. 2881–2890.

19. M. Zhang, W. Li, and Q. Du, "Diverse region-based CNN for hyperspectral image classification," IEEE Transactions on Image Processing, vol. 27, no. 6, pp. 2623–2634, 2018.

20. A. Vaswani, N. Shazeer, N. Parmar, J. Uszkoreit, L. Jones, A. N. Gomez, L. Kaiser, and I. Polosukhin, "Attention is all you need," Advances in NeurIPS, vol. 30, 2017.

21. Dosovitskiy, A., et al.: "An image is worth 16x16 words: Transformers for image recognition at scale." arXiv preprint arXiv:2010.11929.

22. X. Li, F. Xu, R. Xia, T. Li, Z. Chen, X. Wang, Z. Xu, and X. Lyu, "Encoding contextual information by interlacing transformer and convolution for remote sensing imagery semantic segmentation," Remote Sensing, vol. 14, no. 16, p. 4065, 2022.

23. F. Wang, J. Ji, and Y. Wang, "DSViT: Dynamically scalable vision transformer for remote sensing image segmentation and classification," IEEE J-STARS, 2023.

24. Q. Zhao, J. Liu, Y. Li, and H. Zhang, "Semantic segmentation with attention mechanism for remote sensing images," IEEE Transactions on Geoscience and Remote Sensing, vol. 60, pp. 1–13, 2021.

25. Liu, Y., Fan, B., Wang, L., Bai, J., Xiang, S., Pan, C.: Semantic labeling in very high resolution images via a self-cascaded convolutional neural network. ISPRS Journal of Photogrammetry and Remote Sensing, vol. 145, pp. 78–95, 2018.

26. Yue, K., Yang, L., Li, R., Hu, W., Zhang, F., Li, W.: TreeUNet: Adaptive tree convolutional neural networks for subdecimeter aerial image segmentation. ISPRS Journal of Photogrammetry and Remote Sensing, vol. 156, pp. 1–13, 2019.

27. Zheng, Z., Zhong, Y., Wang, J., Ma, A.: Foreground-aware relation network for geospatial object segmentation in high spatial resolution remote sensing imagery. In: Proceedings of the IEEE/CVF Conference on Computer Vision and Pattern Recognition, 2020, pp. 4096–4105.

28. Ma, A., Wang, J., Zhong, Y., Zheng, Z.: FactSeg: Foreground activation-driven small object semantic segmentation in large-scale remote sensing imagery. IEEE Transactions on Geoscience and Remote Sensing, vol. 60, pp. 1–16, 2021.

29. Ding, L., Tang, H., Bruzzone, L.: Lanet: Local attention embedding to improve the semantic segmentation of remote sensing images. IEEE Transactions on Geoscience and Remote Sensing, vol. 59, no. 1, pp. 426–435, 2020.

30. Contributors, P.: PaddleSeg, end-to-end image segmentation kit based on PaddlePaddle, 2019.

31. Fu, J., Liu, J., Tian, H., Li, Y., Bao, Y., Fang, Z., Lu, H.: Dual attention network for scene segmentation. In: IEEE Conference on Computer Vision and Pattern Recognition, CVPR 2019, 2019, pp. 3146–3154.

32. Zhu, Z., Xu, M., Bai, S., Huang, T., Bai, X.: Asymmetric non-local neural networks for semantic segmentation. In: 2019 IEEE/CVF International Conference on Computer Vision, ICCV 2019, 2019, pp. 593–602.

33. Dosovitskiy, A., Beyer, L., Kolesnikov, A., Weissenborn, D., Zhai, X., Unterthiner, T., Dehghani, M., Minderer, M., Heigold, G., Gelly, S., Uszkoreit, J., Houlsby, N.: An image is worth 16x16 words: Transformers for image recognition at scale. CoRR, vol. abs/2010.11929, 2020.

34. Touvron, H., Cord, M., Sablayrolles, A., Synnaeve, G., Jégou, H.: Going deeper with image transformers. In: 2021 IEEE/CVF International Conference on Computer Vision, ICCV 2021, 2021, pp. 32–42.

35. Liu, Z., Lin, Y., Cao, Y., Hu, H., Wei, Y., Zhang, Z., Lin, S., Guo, B.: Swin Transformer: Hierarchical Vision Transformer using Shifted Windows. In: 2021 IEEE/CVF International Conference on Computer Vision, ICCV 2021, 2021, pp. 9992–10002.

36. Yuan, Y., Chen, X., Wang, J.: Object-contextual representations for semantic segmentation. In: Computer Vision - ECCV 2020 - 16th European Conference, 2020, pp. 173–190.

37. Strudel, R., Pinel, R. G., Laptev, I., Schmid, C.: Segmenter: Transformer for semantic segmentation. In: 2021 IEEE/CVF International Conference on Computer Vision, ICCV 2021, 2021, pp. 7242–7252.

38. Xie, E., Wang, W., Yu, Z., Anandkumar, A., Alvarez, J. M., Luo, P.: Segformer: Simple and efficient design for semantic segmentation with transformers. In: Advances in Neural Information Processing Systems 34: Annual Conference on Neural Information Processing Systems 2021, NeurIPS 2021, 2021, pp. 12077–12090.

39. R. Ranftl, A. Bochkovskiy, and V. Koltun, "Vision transformers for dense prediction," in 2021 IEEE/CVF International Conference on Computer Vision, ICCV 2021, Montreal, QC, Canada, October 10-17, 2021. IEEE, 2021, pp. 12 159–12 168.

40. L. Xu, W. Ouyang, M. Bennamoun, F. Boussa¨ıd, and D. Xu, "Multi-class token transformer for weakly supervised semantic segmentation," in IEEE/CVF Conference on Computer Vision and Pattern Recognition, CVPR 2022, New Orleans, LA, USA, June 18-24, 2022. IEEE, 2022, pp. 4300–4309.

41. H. Chen, Z. Qi, and Z. Shi, "Remote sensing image change detection with transformers," IEEE Transactions on Geoscience and Remote Sensing, vol. 60, pp. 1–14, 2021.

42. M. Zhou, J. Huang, Y. Fang, X. Fu, and A. Liu, "Pan-sharpening with customized transformer and invertible neural network," in Proceedings of the AAAI conference on artificial intelligence, vol. 36, no. 3, 2022, pp. 3553–3561.

43. T. Xiao, Y. Liu, Y. Huang, M. Li, and G. Yang, "Enhancing multiscale representations with transformer for remote sensing image semantic segmentation," IEEE Transactions on Geoscience and Remote Sensing, vol. 61, pp. 1–16, 2023.

44. A. Vaswani, N. Shazeer, N. Parmar, J. Uszkoreit, L. Jones, A. N. Gomez, L. Kaiser, and I. Polosukhin, "Attention is all you need," in Advances in Neural Information Processing Systems 30: Annual Conference on Neural Information Processing Systems 2017, December 4-9, 2017, Long Beach, CA, USA, 2017, pp. 5998–6008.

45. H. Touvron, M. Cord, M. Douze, F. Massa, A. Sablayrolles, and H. J´egou, "Training data-efficient image transformers & distillation through attention," in Proceedings of the 38th International Conference on Machine Learning, ICML 2021, 18-24 July 2021, Virtual Event, ser. Proceedings of Machine Learning Research, vol. 139. PMLR, 2021, pp. 10 347–10 357.

46. Ren and Malik, "Learning a classification model for segmentation," in Proceedings Ninth IEEE International Conference on Computer Vision, 2003, pp. 10–17 vol.1.

47. Y. Zhang, B. Pang, and C. Lu, "Semantic segmentation by early region proxy," in IEEE/CVF Conference on Computer Vision and Pattern Recognition, CVPR 2022, New Orleans, LA, USA, June 18-24, 2022. IEEE, 2022, pp. 1248–1258.

48. A. Steiner, A. Kolesnikov, X. Zhai, R. Wightman, J. Uszkoreit, and L. Beyer, "How to train your VIT? Data, augmentation, and regularization in vision transformers," CoRR, vol. abs/2106.10270, 2021.

49. M. Contributors, "MMSegmentation: Openmmlab semantic segmentation toolbox and benchmark," https://github.com/open-mmlab/mmsegmentation, 2020.

50. N. K. Sinha and M. P. Griscik, "A stochastic approximation method," IEEE Trans. Syst. Man Cybern., vol. 1, no. 4, pp. 338–344, 1971.

51. L. Chen et al., "Encoder-decoder with atrous separable convolution for semantic image segmentation," in Proceedings of ECCV 2018, Munich, Germany, September 8-14, 2018, Lecture Notes in Computer Science, vol. 11211, Springer, 2018, pp. 833–851.

52. Z. Zhou et al., "UNet++: Nested U-Net for medical image segmentation," in Proceedings of DLMIA 2018 and ML-CDS 2018, Granada, Spain, September 20, 2018, Lecture Notes in Computer Science, vol. 11045, Springer, 2018, pp. 3–11.

53. H. Zhao, J. Shi, X. Qi, X. Wang, and J. Jia, "Pyramid scene parsing network," in Proceedings of CVPR 2017, Honolulu, HI, USA, July 21-26, 2017, IEEE Computer Society, 2017, pp. 6230–6239.

54. J. Wang et al., "Deep high-resolution representation learning for visual recognition," IEEE Trans. Pattern Anal. Mach. Intell., vol. 43, no. 10, pp. 3349–3364, 2021.

55. Y. Jin, D. Han, and H. Ko, "Trseg: Transformer for semantic segmentation," Pattern Recognition Letters, vol. 148, pp. 29–35, 2021.

56. Q. He et al., "Multimodal remote sensing image segmentation with intuition-inspired hypergraph modeling," IEEE Transactions on Image Processing, vol. 32, pp. 1474–1487, 2023.